\documentclass{article}
\usepackage[colorlinks]{hyperref}
\usepackage{spconf,amsmath,graphicx}
\usepackage{stfloats}
\usepackage{float}

\usepackage{tikz}
\usepackage{comment}
\usepackage{amsmath,amssymb} 
\usepackage{bbm}
\usepackage{booktabs}
\usepackage{graphicx}
\usepackage{subfig}
\usepackage{bm}
\usepackage{multirow}

\makeatletter
\renewcommand{\maketag@@@}[1]{\hbox{\m@th\normalsize\normalfont#1}}%
\makeatother

\title{Hint-dynamic Knowledge Distillation}

\name{Yiyang Liu\textsuperscript{1,2}, Chenxin Li\textsuperscript{1}, Xiaotong Tu\textsuperscript{*}\textsuperscript{1,2}\thanks{The study is supported partly by the National Natural Science Foundation of China under Grants 52105126, 82172033, U19B2031, 82272071, 62271430, and 61971369.(\textsuperscript{*}Corresponding author: Xiaotong Tu, xttu@xmu.edu.cn)}, Xinghao Ding\textsuperscript{1,2}, Yue Huang\textsuperscript{1,2}}
\address{\textsuperscript{1}School of Informatics, Xiamen University, China \\
\textsuperscript{2}Institue of Artificial Intelligent, Xiamen University, China \\
}

\begin{document}
\begin{sloppypar}
%
\maketitle
\begin{abstract}
Knowledge Distillation (KD) transfers the knowledge from a high-capacity teacher model to promote a smaller student model.
Existing efforts guide the distillation by matching their prediction logits, feature embedding, \textit{etc.}, while leaving how to efficiently utilize them in junction less explored.
In this paper, we propose \textbf{H}int-dynamic \textbf{K}nowledge \textbf{D}istillation, dubbed HKD, which excavates the knowledge from the teacher's hints in a dynamic scheme.
The guidance effect from the knowledge hints usually varies in different instances and learning stages, which motivates us to customize a specific hint-learning manner for each instance adaptively.
Specifically, a meta-weight network is introduced to generate the instance-wise weight coefficients about knowledge hints in the perception of the dynamical learning progress of the student model.
We further present a weight ensembling strategy to eliminate the potential bias of coefficient estimation by exploiting the historical statics. 
Experiments on standard benchmarks of CIFAR-100 and Tiny-ImageNet manifest that the proposed HKD well boost the effect of knowledge distillation tasks. 
\end{abstract}
\begin{keywords}
Knowledge Distillation; Dynamic Network; Meta-Learning
\end{keywords}

\section{Introduction}
\label{sec:intro}
Whilst deep neural networks (DNNs) have achieved remarkable success in computer vision, most of these well-performed models are difficult to deploy on edge devices in practical scenarios due to the high computational costs. 
To alleviate this, light-weight DNNs have been investigated a lot. 
The typical approaches mainly include parameter quantization \cite{gong2014compressing, wu2016quantized, lin2020hrank}, network pruning \cite{han2015learning, yamamoto2021learnable, liu2017learning}, knowledge distillation (KD) \cite{hinton2015distilling}, \textit{etc.}
Among them, the KD topic has gained increasing popularity in various vision tasks due to its simplicity to be integrated into other model compression pipelines.

\begin{figure}[t]
\centering
\setlength{\abovecaptionskip}{0.2cm}
\includegraphics[width=1\linewidth]{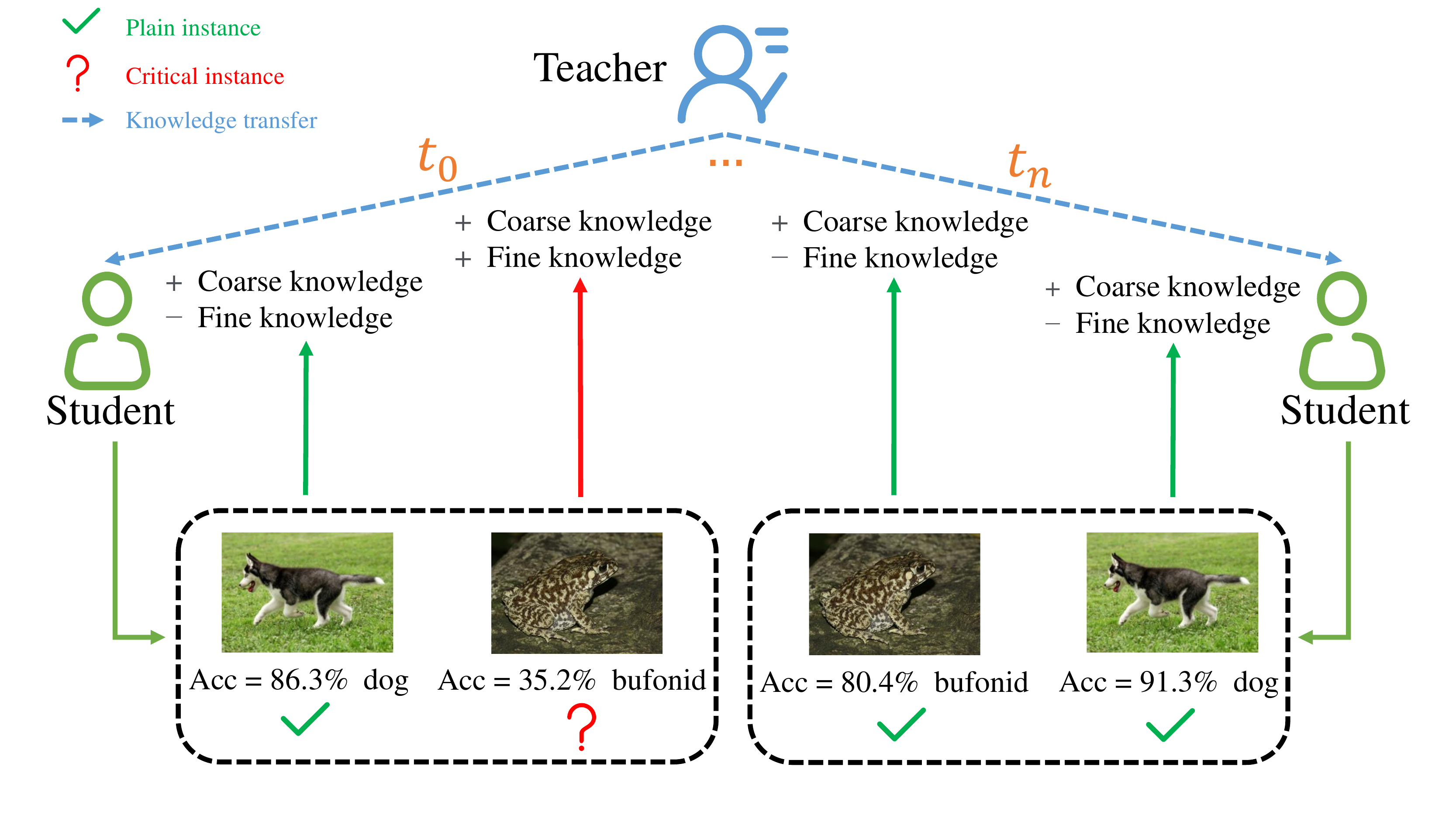}
\caption{
Motivation of the proposed HKD. 
Unlike the existing KD paradigm that exploits different knowledge hints in a pre-defined fashion, 
our HKD adaptively customizes the learning fashion on each instance at different training iteration $t$.  
}
\label{fig1}
\vspace{-1.5em}
\end{figure}

The core idea of KD \cite{hinton2015distilling} is to distill the knowledge from the cumbersome teacher model to strengthen the compressed student by matching their posterior distribution of class labels as knowledge hints.
Numerous subsequent works further explore various new forms of matching hints, such as intermediate representation \cite{romero2014fitnets,chen2021distilling, aguilar2020knowledge}, attention maps \cite{zagoruyko2016paying}, mutual information~\cite{ahn2019variational, peng2019correlation, tung2019similarity}, structural knowledge \cite{park2019relational,zhu2021complementary, tian2019contrastive}, \textit{etc.}
By extracting knowledge from the soft labels as coarse distillation, most of these methods further leverage more fine-grained distillation from the explored novel knowledge hints. 
Concretely, they combine the guidance from different grains with a pre-defined combination coefficient, considering the teacher's knowledge hints arguably keeps consistent distillation value across different instances throughout the training process.
Nevertheless, the dynamical capacity of the student model is neglected by most of these methods.
In this regard, the current KD paradigm tends to fail in modeling and perceiving the dynamical distillation effect of different knowledge hints.

\begin{figure*}[ht]
\centering
\setlength{\abovecaptionskip}{0cm}
\includegraphics[width=0.95\linewidth]{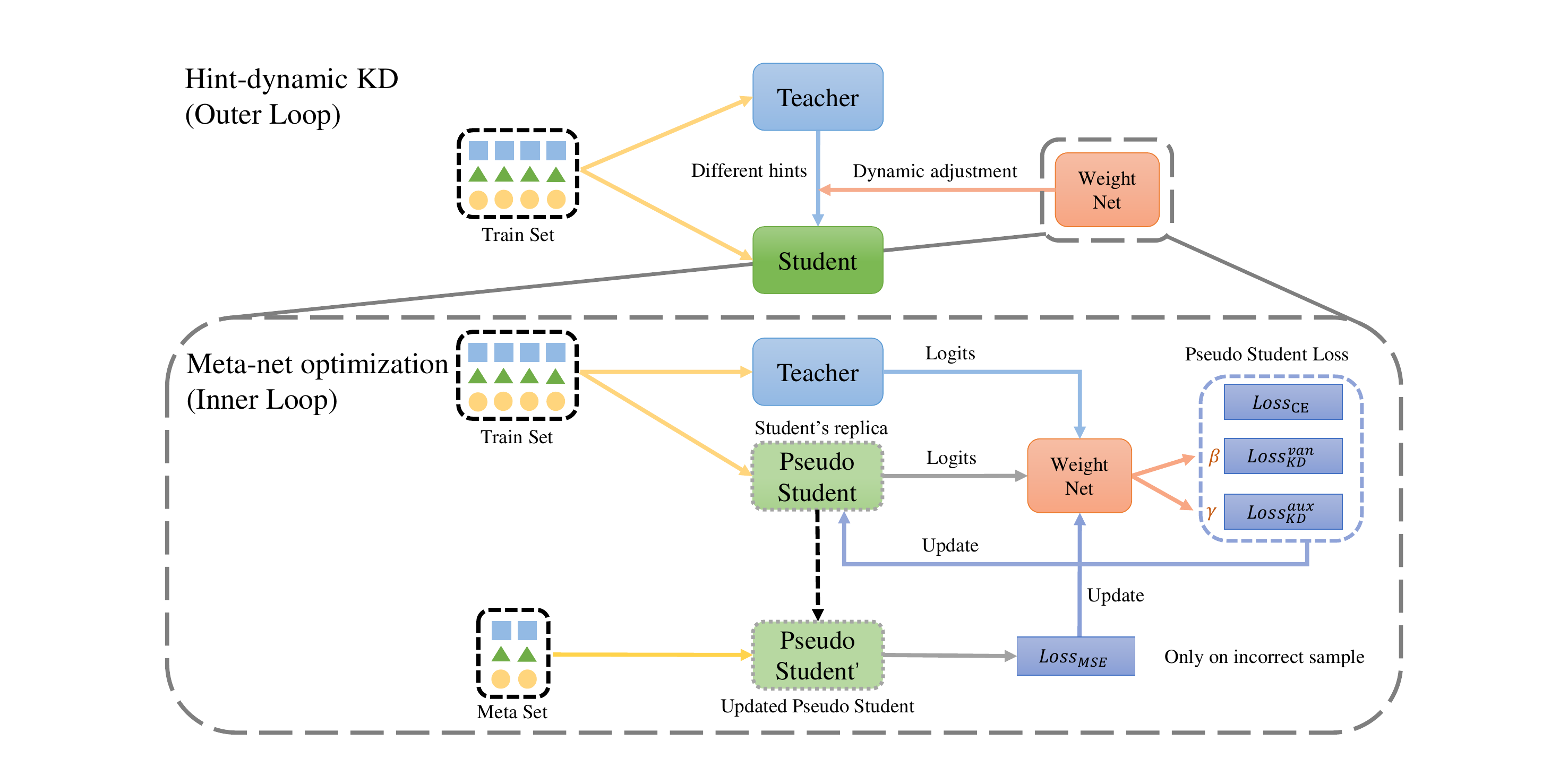}
\caption{
Overview of the proposed HKD. 
The optimal learning fashion for different knowledge hints on each instance is dynamically estimated based on meta-learning. 
A meta-weight network in the perception of the feedback from student estimates the learning coefficients for different knowledge hints. 
The KD pipeline and the introduced meta-weight network are optimized jointly in a nested loop optimization.
}

\label{fig2}
\end{figure*}

To ameliorate the above issue, we present a novel hint-dynamic scheme from the insight of efficiently utilizing diverse knowledge representation.
Fig.\,\ref{fig1} depicts the motivation of our proposed framework.
The learning progress of the student differs in instances across the distillation procedure. 
For the plain instances which are certain for the student, simply the coarse knowledge from the rudimentary soft labels is enough to guide the distillation.
In contrast, for those critical ones who are not well-learned, more fine-grained knowledge from other hints like feature embedding is introduced.

Our insight is to dynamically generate a customized learning fashion to handle different knowledge hints according to the aptitude of student. 
To this end, we formulate the importance of each knowledge hint as a variable dependent on the input instance and model the learning fashion as weight coefficients of the KD loss from different hints.
A meta-weight network is further leveraged, where an inner loop is set up to train the meta-weight network to generate weights for each instance, while an outer loop is further introduced to guide the efficient KD by the updated meta-weight.
To alleviate the estimation bias of optimal weights, we further propose a weight ensembling strategy utilizing historical statics.

Our contribution can be summarized as:

\begin{itemize}
\item{We propose a novel Hint-dynamic Knowledge Distillation framework that enables dynamic learning for various knowledge hints adaptively.}
\item{We introduce a meta-learning based method that dynamically assigns the weight coefficients about distillation loss for each sample.}
\item{We derive an uncertainty-based weight ensembling strategy, which alleviates the adverse effect of the unreliable estimation of meta-weight via historical statics.}
\item{Experiments demonstrate the superior performance of the proposed HKD on benchmark datasets.}
\end{itemize}

\section{Method}
\label{method}

\subsection{Preliminaries}
\label{pre}
In the task of KD, given a pre-trained teacher model $\mathcal{T}$ and a student model $\mathcal{S}$ on a training data set $\mathcal{X}$, 
the KL divergence between the student output $p_\mathcal{S}(x)$ and $p_\mathcal{T}(x)$ is minimized as the vanilla KD version \cite{hinton2015distilling}: 
\begin{equation}
  \label{eq_kl}
  \mathcal{L}_{KD}^{van}=\sum_{x\in\mathcal{X}}p_\mathcal{T}(x) \cdot log\frac{p_{\mathcal{T}}(x)}{p_{\mathcal{S}}(x)}
\end{equation}
Subsequently, the community further explores a extensive variety of hint forms for knowledge transfer beyond the prediction labels, like intermediate layers \cite{romero2014fitnets}, attention maps \cite{zagoruyko2016paying}, \textit{etc.}
Specifically, they leverage an auxiliary guidance signals from the teacher by using the matching loss $\mathcal{L}_{KD}^{aux}$ for the exploited hints, which is utilized to update the student $\mathcal{S}$ over training data set $\mathcal{X}$:
\begin{equation}
  \label{eq_lossall}
  \mathcal{L}({\mathcal{S};\mathcal{X}})=\sum_{x\in\mathcal{X}} \mathcal{L}_{CE}(x) + \beta\mathcal{L}_{KD}^{van}(x) + \gamma\mathcal{L}_{KD}^{aux}(x)
\end{equation}
where $\mathcal{L}_{CE}$ is the cross entropy loss whereas $\beta,\gamma$ are the weight coefficients for different distillation losses of each sample, respectively, which are designed empirically to keep fixed for all the instances across the whole training procedure in conventional KD methods, despite the different learning progress of the student model for each sample.
As a core distinction, we propose to dynamically adjust the guidance manner from the teacher, which emphasizes the varying demand for different knowledge hints for each instance at different iterations.

\subsection{Hint-dynamic Knowledge Distillation}
\label{training}
\noindent\textbf{Overview.}\ 
Towards the goal of the dynamic distillation scheme which adaptively allocates the hint weights for each instance as the training procedure goes, one naive solution is to utilize an uncertainty-based metric \cite{li2021dynamic} to evaluate the instance-wise learning progress, which is somehow unreliable.
To ameliorate this issue, we leverage the merits of meta-learning, which provide a second-order optimization framework to alleviate the estimation bias \textit{w.r.t.} learning degree. 
Specifically, a meta-weight network (meta-net) is introduced to explicitly encode the importance of each instance as well as generate the dynamic estimation for the subsequent distillation manner.
By utilizing the inner and outer loop, the introduced meta-net and the teacher-student distillation framework promote each other in our proposed scheme. Fig.\,\ref{fig2} depicts the workflow of the proposed HKD.
\vspace{0.5em}

\noindent\textbf{Meta-Weight Network.}\ 
To perform the instance-wise dynamic estimation of the optimal learning fashion \textit{w.r.t.} different knowledge hints, we design a meta-net $\mathcal{W}$ to generate the weight coefficients for each instance prior to every learning iteration of the student. 
Holding the insight that the cross-model relation matters, we feed not only the prediction logits of the student but also the teacher's prediction into meta-net, in such way the weight for the sample $x$ can be written as:
\begin{equation}
  \label{eq_generate_weight} 
  \beta(x),\gamma(x) = \mathcal{W}(p_\mathcal{S}(x),p_\mathcal{T}(x))
\end{equation}
Practically, this meta-net is easy to implement, \textit{e.g.}, a 2-layer Multi-Layer Perceptron (MLP) with a given weight range.
In what follows, we further leverage a technique of pseudo student generation to perform the inner-loop optimization for our HKD, which is inspired by the insight of meta-learning \cite{finn2017maml}.
\vspace{0.5em}

\noindent\textbf{Inner Loop via Pseudo Student Generation.}\ \ 
In the utilization of a meta-net, we exploit a technique to update a pseudo copy of student model to make perception of the model performance using the meta-net.
Specifically, a meta-set $\mathcal{X}_{meta}$ is held out from the whole training data set $\mathcal{X}$, \textit{i.e.}, $|\mathcal{X}_{meta}| \ll |\mathcal{X}|$,
and we perform a one-step gradient update for the student as a pseudo version $\mathcal{S}_{p}$:
\begin{small}
\begin{equation}
\label{eq_loss_sp}
  \mathcal{L}({\mathcal{S}_p};\mathcal{X})=
  \sum_{x\in\mathcal{X}}\mathcal{L}_{CE}(x) + \beta(x)\mathcal{L}_{KD}^{van}(x) + \gamma(x)\mathcal{L}_{KD}^{aux}(x)
\end{equation}
\end{small}
The mean-square error (MSE) of pseudo student on the meta-set reflects the quality of weight estimation of the meta-net, which can be further utilized to optimize the meta-net $\mathcal{W}$:
\begin{equation}
  \label{eq_loss_metanet}
  \mathcal{L}(\mathcal{W};\mathcal{X}_{meta}^{er})= 
  \sum_{x\in\mathcal{X}_{meta}^{er}}
  \mathcal{L}_{MSE}(p_{S_\mathcal{P}}(x),GT(x))
\end{equation}
where $\mathcal{X}_{meta}^{er}$ denotes the incorrect results of pseudo student's output in the meta-set \cite{liu2022meta}, 
$p_{S_\mathcal{P}}(x)$ is the prediction probability of the pseudo student, and $GT(x)$ returns the ground-truth probability value of the corresponding sample.
\vspace{0.5em}

\noindent\textbf{Outer Loop Optimization via Second-order Gradient.}\ 
To facilitate the guidance effect of the updated meta-net in the inner loop, we further leverage an outer loop process, in which the student model acquires the knowledge from the teacher according to the generated fashion when dealing with the knowledge hints.
In this regard, a standard knowledge distillation process is introduced as Eq.\,\ref{eq_lossall}.
Then, by iteratively executing the two preceding loops, we can formulate a nested optimization problem:
\begin{equation}
  \begin{aligned}
    & \mathop{min}\limits_{\mathcal{S}}\quad \mathcal{L}({\mathcal{S}};\mathcal{X}) \\
    & s.t.\quad \mathcal{W}=\mathop{argmin}\limits_{\mathcal{W}}\ \mathcal{L}(\mathcal{W};\mathcal{X}_{meta}^{er})
\end{aligned}
\end{equation}
where the outer loop is formulated as a problem to search for the optimal hint weights while constrained by the inner loop.

\subsection{Meta-Weight Ensembling}
\label{mwe}
The effect of the proposed meta-learning based framework relies on the accurate estimation of the hint coefficients, \textit{i.e.}, the output of the meta-net, whereas the transient state of this weight is not always reliable.
To address this issue, we further propose a strategy to generate a more robust hint weights via the temporary ensembling:
\begin{equation}
\resizebox{.91\hsize}{!}{$
(\beta^{t}(x),\gamma^{t}(x))=
\left\{
\begin{aligned}
	\epsilon \cdot (\beta^{t-1}(x),\gamma^{t-1}(x)) +
	(1-\epsilon) \cdot &(\beta(x),\gamma(x))  &u(x)<u_{th} \\
	&(\beta(x),\gamma(x))  &u(x)\geq u_{th}
\end{aligned}
\right.\\
$}
\label{eq_uner}
\end{equation}
where $t$ denotes the current step and $t-1$ is the previous step.
$\epsilon$ controls the ratio between the weight of $t$ and $t-1$. 
$u_{th}$ is the threshold value of uncertainty.
This mechanism is applied to the updating of student in the outer loop, which means $\beta_{x}^{t}, \gamma_{x}^{t}$ are calculated by Eq.\,\ref{eq_uner}. $u(x)$ representing the uncertainty of sample $x$, can be modeled by the prediction entropy: $u(x) = -{p_\mathcal{S}(x)log(p_\mathcal{S}(x)})$.

\section{Experiments}
\label{exp}

\noindent\textbf{Datasets.}\ 
Experiments are conducted on the benchmark datasets of CIFAR-100 \cite{krizhevsky2009learning} and Tiny-ImageNet \cite{le2015tiny}. 
CIFAR-100 contains 50K 32$\times$32 training images with 500 images per class and 10K test images with 100 images per class. Tiny-ImageNet is a subset version of ImageNet with 200 classes, where each image is down-sampled to 64$\times$64. The images in each class are split as 500/50 for training and testing, respectively.
\vspace{0.5em}

\noindent\textbf{Implementation Details.}\ \ 
Following the common practice in KD \cite{tian2019contrastive, song2022spot}, we set the total number of training epochs to 240 while the batch size to 64. 
We use stochastic gradient descent (SGD) as the optimizer for the student model. 
Except for ShuffleNet V1, which is set to 0.01, the initial learning rate is 0.05. Weight decay is set to $5\times10^{-4}$ , 
For meta-net, we adopt an Adam optimizer with the initial learning rate $1\times10^{-3}$. For the meta-set, we set size to 1000 on CIFAR-100 and 2000 on Tiny-ImageNet considering 10 samples per class. 
The training interval of the inner loop is 100. 
We search for the optimal dynamic weights $\beta$ and $\gamma$ with a searching range $\l=0.5$ around the initial value $1$.
For the weight ensembling, an uncertainty threshold $u_{th}$ of 0.6 is adopted, and $\epsilon$ is 0.5. 

\subsection{Comparisons with State-of-the-art Methods}  
\noindent\textbf{Results on CIFAR-100.}\ 
We test the performance of our method when combined with three state-of-the-art KD works, including Fitnet \cite{romero2014fitnets}, VID \cite{ahn2019variational} and CRD \cite{tian2019contrastive}. 
We directly cite the quantitative results reported in their papers \cite{tian2019contrastive}. 
The results are shown in Tab.\,\ref{CIFAR-100acc}.
\textit{Teacher} and \textit{Student} denote the accuracy of the teacher and student models when they are trained individually.  
We can see that combining the proposed HKD with the modern KD methods leads to a significant improvement. 
Besides, we compare with two other adaptive distillation works \cite{li2021dynamic, song2022spot}, and it can be seen that HKD achieves better results in most of the experiments.
\vspace{0.5em}

\noindent\textbf{Results on Tiny-ImageNet.}\ 
Following KD on Tiny-ImageNet common practice \cite{song2022spot}, experiments are conducted with Vgg13 → Vgg8, WRN\_40\_2 → WRN\_16\_2 and ResNet110 → ResNet20. Tab.\,\ref{tinyimagenetacc} presents the results, which indicate that HKD continues to outperform other works.
\begin{table}[!t]
    \begin{minipage}{1.0\linewidth}
    \renewcommand{\arraystretch}{1.3}
    \setlength{\abovecaptionskip}{5pt}
    \caption{Top-1 Test Acc.\,(\%) of the student networks on CIFAR-100.}
    \label{CIFAR-100acc}
    \resizebox{1\linewidth}{!}{
    \begin{tabular}{c|cccc|cccc|cccc}
        \hline
        \multicolumn{1}{l|}{} & \multicolumn{4}{c|}{ResNet32x4 → ResNet32} & \multicolumn{4}{c|}{WRN\_40\_2 → ShuffleNetV1} & \multicolumn{4}{c}{ResNet50 → Vgg8} \\ \hline
        \multicolumn{1}{l|}{} & Std & D-KD & S-KD & Ours & Std & D-KD & S-KD & Ours & Std & D-KD & S-KD & Ours \\
        Teacher & \multicolumn{1}{l}{74.92} & - & - & - & \multicolumn{1}{l}{75.61} & - & - & - & \multicolumn{1}{l}{79.34} & - & - & - \\
        Student & \multicolumn{1}{l}{71.14} & - & - & - & \multicolumn{1}{l}{70.50} & - & - & - & \multicolumn{1}{l}{70.36} & - & - & - \\ \hline
        Fitnet & 72.35 & 72.63 & 72.63 & \textbf{72.83} & 75.67 & 75.55 & 75.93 & \textbf{76.25} & 73.24 & 73.3 & \textbf{73.75} & \textbf{73.75} \\
        VID & 72.29 & 72.59 & 72.85 & \textbf{72.94} & 75.88 & 76.12 & 76.21 & \textbf{76.41} & 73.46 & 73.57 & \textbf{73.92} & 73.79 \\
        CRD & 72.99 & 73.26 & 73.21 & \textbf{73.78} & 76.03 & 76.2 & 76.26 & \textbf{76.82} & 74.58 & 73.94 & 74.81 & \textbf{74.96} \\ \hline
    \end{tabular}}
    \end{minipage}
    \\[8pt]  
    \begin{minipage}{1\linewidth}
    \renewcommand{\arraystretch}{1.3}
    \setlength{\abovecaptionskip}{5pt}
    \caption{Top-1 Test Acc.\,(\%) of the student networks on Tiny-ImageNet.}
    \label{tinyimagenetacc}
    \resizebox{1.0\linewidth}{!}{
    \begin{tabular}{c|cccc|cccc|cccc}
        \hline
        \multicolumn{1}{l|}{} & \multicolumn{4}{c|}{Vgg13 → Vgg8} & \multicolumn{4}{c|}{WRN\_40\_2 → WRN\_16\_2} & \multicolumn{4}{c}{ResNet110 → ResNet20} \\ \hline
        \multicolumn{1}{l|}{} & Std & D-KD & S-KD & Ours & Std & D-KD & S-KD & Ours & Std & D-KD & S-KD & Ours \\
        T & \multicolumn{1}{l}{60.09} & - & - & - & \multicolumn{1}{l}{61.26} & - & - & - & \multicolumn{1}{l}{58.46} & - & - & - \\
        S & \multicolumn{1}{l}{56.03} & - & - & - & \multicolumn{1}{l}{57.17} & - & - & - & \multicolumn{1}{l}{51.89} & - & - & - \\ \hline
        Fitnet & 58.33 & 58.88 & 59.10 & \textbf{59.67} & 58.88 & 58.96 & 59.33 & \textbf{59.55} & 54.04 & 54.10 & 54.25 & \textbf{54.36} \\
        VID & 58.55 & 58.84 & 58.80 & \textbf{59.05} & 58.78 & 58.85 & 58.99 & \textbf{59.15} & 53.94 & 54.12 & 54.28 & \textbf{54.42} \\
        CRD & 58.88 & 59.65 & 59.38 & \textbf{59.51} & 59.42 & 59.64 & 59.87 & \textbf{60.22} & 54.69 & 54.99 & 55.28 & \textbf{55.65} \\ \hline
        \end{tabular}}
    \end{minipage}
\end{table}

\subsection{Further Empirical Analysis}
\noindent\textbf{Ablation Study.}\ 
We conduct ablation studies on CIFAR-100 to validate the effect of each component in our HKD.
The results are shown in Tab.\,\ref{table3}. 
Note that the experiments here are based on the feature hints from CRD \cite{tian2019contrastive}. 
(1) \textit{Static} denotes the baseline which adopts a fixed weight of knowledge hints during the whole training procedure.
(2) \textit{Un-Dy} models the dynamic property \textit{w.r.t.} the knowledge hints using a uncertainty-based approach as in \cite{li2021dynamic}.
We can see the performance gain brought by utilizing the dynamic modeling for the weight.
(3) \textit{MWN} further exploits the meta-weight network to generate the dynamic weight in different searching range $\l$ of 0.5 and 1 around the initial value.
It can be seen that an appropriate searching range leads to a better distillation effect.
(4) When equipped with meta-weight ensembling strategy as \textit{MWE} to facilitate a more robust estimation of dynamic weight, the accuracy of the student model distilled by our full HKD method achieves the best accuracy, which indicates each component of our HKD plays its own role.

\begin{table}[t]
  \centering
\setlength{\abovecaptionskip}{5pt}
  \caption{Top-1 Test Acc.\,(\%) of ablation studies on CIFAR-100.}
  \resizebox{1\linewidth}{!}{
    \setlength{\parindent}{0pt}
    \setlength{\tabcolsep}{2pt}{
    \begin{tabular}{c|cc|cc|c}
    \hline
     & Static & Un-Dy & MWN\,($\l=1$) & MWN\,($\l=0.5$) & HWN+MWE (full) \\ \hline
    R32x4→R32 & 72.99 & 73.26 & 73.30 & 73.37 & 73.78 \\
    W40-2→SN1 & 76.03 & 76.20 & 76.28 & 76.46 & 76.82 \\ \hline
    \end{tabular}
    }
  }
  \label{table3}%
  \vspace{-0.1cm}
\end{table}

\begin{figure}[h]
\begin{minipage}[b]{0.49\linewidth}
  \centering
  \centerline{\includegraphics[width=1\linewidth,keepaspectratio]{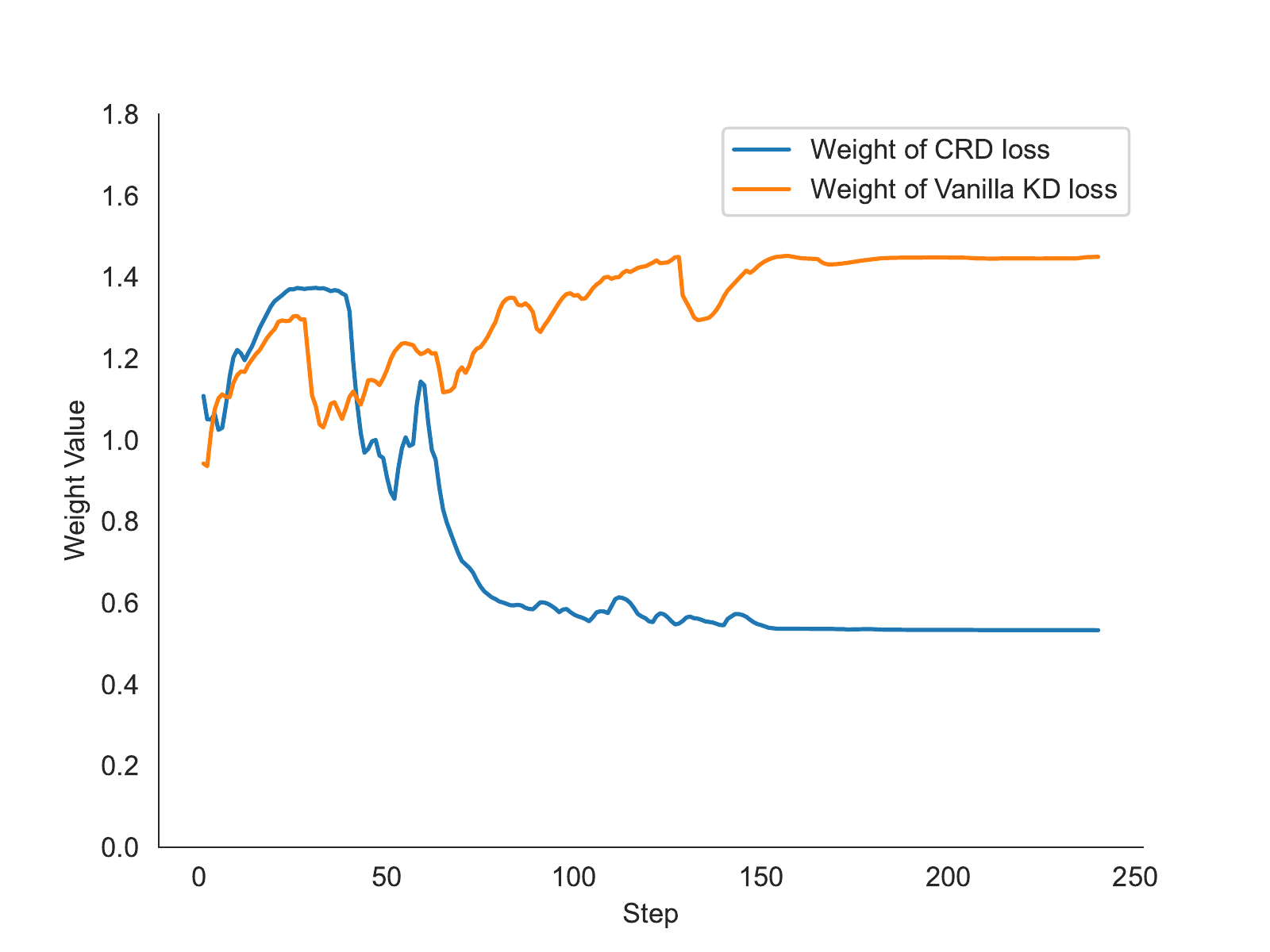}}
  \centerline{(a) Vgg13 → Vgg8}\medskip
\end{minipage}
\hfill
\begin{minipage}[b]{0.49\linewidth}
  \centering
  \centerline{\includegraphics[width=1\linewidth,keepaspectratio]{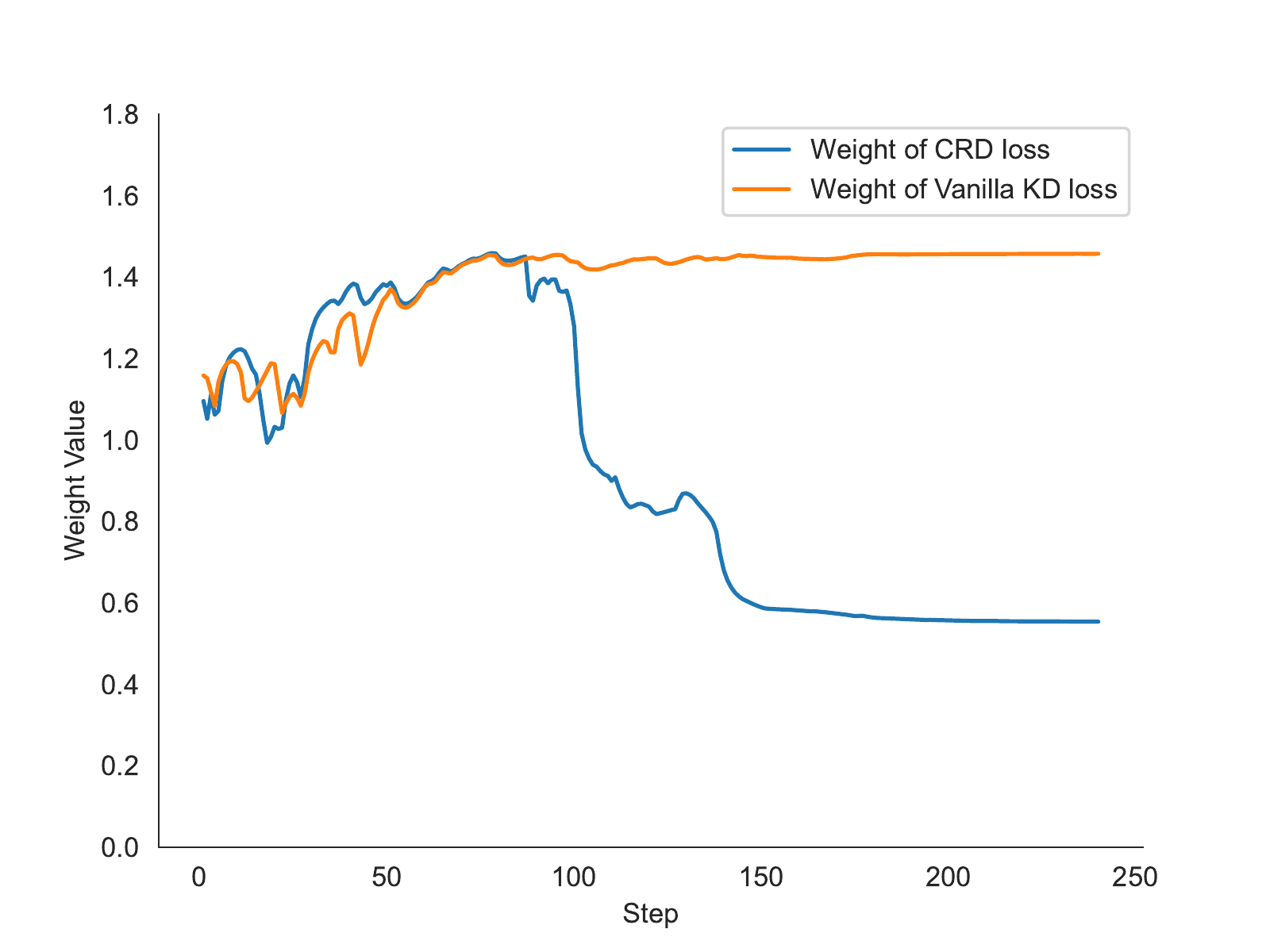}}
  \centerline{(b) WRN\_40\_2 → WRN\_16\_2}\medskip
\end{minipage}
\caption{The curves of the learned hint weights in two experiments on Tiny-ImageNet.}
\label{fig3}
\vspace{-1.7em}
\end{figure}

\vspace{0.5cm}
\noindent\textbf{Visualization on Meta-weight Estimation.}\ 
Fig.\,\ref{fig3} shows the curves of the batch-wise average of weight variation on Tiny-Imagenet.
It can be observed that most samples are uncertain to the student during the initial stage of distillation, thus the vanilla KD loss and CRD loss weights are relatively large. As the distillation goes on, the student's capacity increases, which requires more guidance from vanilla KD. Consequently, the weight of vanilla KD loss increases while the weight of CRD loss decreases. Finally, the student tends to choose a stable learning fashion compared with the initial state. 

\vspace{-1em}

\section{Conclusion}
\label{conclu}
This paper proposes Hint-dynamic Knowledge Distillation (HKD) to promote knowledge transfer in a dynamic and active fashion.
Instead of using the fixed weight coefficients for knowledge hints from the teacher, we dynamically customize instance-wise learning to facilitate the distillation process of the student.
Specifically, a meta-weight network is leveraged to generate the estimation of optimal learning manner \textit{w.r.t.} knowledge hints, further in the utilization of a meta-learning based optimization framework.
To alleviate the bias of weight estimation, we further explore a strategy of meta-weight ensembling, which adaptively ensemble the hint weights considering the historical statics.
Extensive experiments on benchmark datasets show that HKD outperforms state-of-the-art and off-the-shelf distillation methods.

\clearpage

\bibliographystyle{IEEEbib}
\bibliography{refs}

\end{sloppypar}
\end{document}